%% file: main.tex
\documentclass[sigconf]{acmart}
\usepackage{comment}
\usepackage{amsmath}
\usepackage{graphicx}
\usepackage{enumitem}
\setlist[itemize]{leftmargin=*}

\setcopyright{none}

\AtBeginDocument{%
  \providecommand\BibTeX{{%
    \normalfont B\kern-0.5em{\scshape i\kern-0.25em b}\kern-0.8em\TeX}}}

\copyrightyear{2024}
\acmYear{2024}
\setcopyright{acmlicensed}\acmConference[CODS-COMAD 2024]{7th Joint
International Conference on Data Science \& Management of Data (11th ACM
IKDD CODS and 29th COMAD)}{January 4--7, 2024}{Bangalore, India}
\acmBooktitle{7th Joint International Conference on Data Science \&
Management of Data (11th ACM IKDD CODS and 29th COMAD) (CODS-COMAD
2024), January 4--7, 2024, Bangalore, India}
\acmPrice{15.00}
\acmDOI{10.1145/3632410.3632456}
\acmISBN{979-8-4007-1634-8/24/01}

\begin{document}

\title{Comparative Analysis of Transformers for Modeling Tabular Data: A Casestudy using Industry Scale Dataset}

\author{Usneek Singh$^{1}$, Piyush Arora$^{2}$, Shamika Ganesan$^{2}$, Mohit Kumar$^{2}$, Siddhant Kulkarni$^{2}$, Salil R. Joshi$^{2}$}
  \affiliation{
    \institution{$^{1}$BITS Pilani, India}
    \country{}
    }
  \affiliation{
    \institution{$^{2}$American Express AI Labs, India}
    \country{}
    }
\email{usneeksingh1@gmail.com, {piyush.arora1, shamika.ganesan, mohit.kumar30, siddhant.r.kulkarni1, salilrajeev.joshi}@aexp.com}

\input{abstract}

\maketitle

\input{introduction}
\input{related-work}
\input{methodology}

\input{fraud}

\input{cdss}
\input{lessons-learnt}
\input{conclusion}

\bibliographystyle{ACM-Reference-Format}
\bibliography{ref}
\clearpage

\input{appendix}

\end{document}

%% file: abstract.tex
\begin{abstract}
We perform a comparative analysis of transformer-based models designed for modeling tabular data, specifically on an industry-scale dataset. While earlier studies demonstrated promising outcomes on smaller public or synthetic datasets, the effectiveness did not extend to larger industry-scale datasets. The challenges identified include handling high-dimensional data, the necessity for efficient pre-processing of categorical and numerical features, and addressing substantial computational requirements.

To overcome the identified challenges, the study conducts an extensive examination of various transformer-based models using both synthetic datasets and the default prediction Kaggle dataset (2022) from American Express. The paper presents crucial insights into optimal data pre-processing, compares pre-training and direct supervised learning methods, discusses strategies for managing categorical and numerical features, and highlights trade-offs between computational resources and performance. Focusing on temporal financial data modeling, the research aims to facilitate the systematic development and deployment of transformer-based models in real-world scenarios, emphasizing scalability.

{\it Keywords - transformers, tabular datasets, financial modeling}

\end{abstract}

%% file: introduction.tex
\section{Introduction}
The Transformer model, introduced by Vaswani et al. (2017),
has significantly impacted various domains, including Natural Language Processing (NLP), Computer Vision (CV), and audio processing. Built upon an encoder-decoder structure with an attention mechanism, the Transformer architecture proves adept at capturing sequence dependencies and learning hierarchical representations, making it effective for modeling sequential patterns.
Currently, gradient boosted decision trees~\cite{chen2016xgboost} are considered the state-of-the-art models for financial modeling. Although these models are intuitive, they come with certain limitations:. \\
\hspace{1cm} 1. reliance on manually derived features when dealing with continuous streaming 
data such as frequency features (e.g. how many months), statistical features (e.g. mean of spending)\\
\hspace{1cm} 2. do not effectively capture the dynamic relationships within time series data and instead rely on adhoc combinations of features such as minimum, maximum, sum, average, etc.

Transformers have gained attraction in the tabular domain due to their ability to generate contextual embeddings effectively and handle the limitations mentioned above as compared to tree based models~\citep{huang2019dsanet,huang2020tabtransformer,liu2021gated}, which we explore for modelling financial datasets.

Prior work cites synthetically generated data~\citep{padhi2021tabular,luetto2023one}, which differs significantly from real-world datasets. A significant challenge arises from the actual dimensionality of the datasets - real-world datasets exhibit 200-500 dimensions, whereas synthetic datasets only account for 10-15 dimensions~\citep{luetto2023one}. Additionally, real-world financial datasets frequently suffer from noise such as out-of-range or missing values.

In this study, we leverage various transformer techniques to effectively model financial tabular series data. Our investigation involves comparison and analysis of different i) input processing techniques, ii) model architectures, and iii) training strategies for handling financial tabular data using transformers. 

Specifically, our contributions are: 
\begin{itemize}
    \item Thorough analysis of different transformer architectures for tabular data based on different dimensions such as data preprocessing, training strategies etc.
    \item Comparative analysis of different transformer architectures for synthetic as well as finance industry datasets capturing real-world problems.
\end{itemize}

The paper is organized as follows: Section~\ref{sec:related-work} presents related work on financial modeling \& recent transformers based architectures, Section~\ref{sec:methodology} provides an overview of architectures that we explore in this work. Section~\ref{sec:fraud} presents our results on synthetic dataset, Section~\ref{sec:cdss} showcases our results on industry based dataset. Section~\ref{subsect:generalization} talks about our findings in alternate regression tasks. Section~\ref{sec:lessons} discusses lessons learned and Section~\ref{sec:conclusion} presents conclusion.

%% file: related-work.tex
\section{Related Work}
\label{sec:related-work}

{\bf Traditional approaches for handling tabular data: }
Traditional machine learning models for handling financial tabular data include gradient boosted decision trees (GBDT) and recurrent neural network (RNN) models. GBDT, including XGBoost \cite{chen2016xgboost}, LightGBM \cite{ke2017lightgbm}, and CatBoost \cite{dorogush2018catboost}, are commonly used for managing tabular data. To address the non-differentiability of decision trees, alternative methods with smooth decision functions have been proposed \cite{popov2019neural,hazimeh2020tree}.
Other techniques, such as Factorization Machines \cite{guo2017deepfm}, provide insights into handling tabular data with regularization methods for deep learning models or simple multi-layer perceptron models \cite{kadra2021well}. However, these methods do not effectively capture sequential information and are less suitable for managing data with sequential tabular information. To address this, RNN-based models for handling tabular series, which capture temporal information in a limited sense contrasted to Transformer based approaches have been proposed \cite{ke2018tabnn,alvi2021deep,du2021tabularnet}.

{\bf Transformer based approaches for handling tabular data: } 
Various transformer-based approaches have been developed for handling tabular data, each with different input processing, model architectures, and training mechanisms. 
TabTransformer \cite{huang2020tabtransformer} utilizes a simple transformer model trained with Masked Language Modeling to generate contextual embeddings for categorical and numerical values. SAINT \cite{somepalli2021saint} modifies the transformer block to create embeddings at the single-row level for both categorical and numerical values. 
TabBERT \cite{padhi2021tabular} introduces a hierarchical transformer that captures the hierarchy in sequence data by converting numerical values into categorical form.
Liu et al. \cite{liu2021gated} propose a supervised learning mechanism with two parallel towers to capture attention across time and dimension, presenting a direct approach without intermediate embeddings. Han et al. \cite{han2022luna} develop a framework that separately handles categorical and numerical values using a pre-trained transformer with a joint loss for each category.
TARNet \cite{chowdhury2022tarnet} incorporates knowledge from the end task by alternating between masked language modeling and supervised downstream tasks during pre-training. Crossformer \cite{zhang2023crossformer} captures attention across both dimensions similar to Liu et al., and its architecture resembles that of TabBERT. 
Each transformer-based approach outlined above possesses its own advantages and disadvantages, catering to different requirements and exhibiting varying performance in handling tabular data. Most of these approaches are tried out on synthetic tabular data alone. 

\vspace{-0.2cm}

%% file: methodology.tex
\section{Methodology}
\label{sec:methodology}
In our exploration of transformer techniques for tabular series data, we carefully selected architectures to align with our research objectives. We categorized our training techniques into two main groups: direct supervised training and decoupled pre-training \& fine-tuning (described in section 3.2).
For direct supervised training, we considered CrossFormer \cite{zhang2023crossformer}, Gated Transformer (Twin Tower) \cite{liu2021gated}, and TARNet \cite{chowdhury2022tarnet}. We opted for Twin Tower due to its accessibility and robust performance.
Within the category of decoupled pre-training \& fine-tuning strategies, we explored various models, including TabBERT \cite{padhi2021tabular}, LUNA \cite{han2022luna}, and UniTTab \cite{luetto2023one}. TabBERT \cite{padhi2021tabular} became our standard choice due to its widespread recognition and effectiveness in handling tabular series data. While UniTTab \cite{luetto2023one} and LUNA \cite{han2022luna} share similar techniques with TabBERT, we retained LUNA for its unique approach to enhancing numerical reasoning within language models, a crucial aspect of our research.
While our study didn't encompass all available models, our selection was deliberate, aiming to provide representation from each category.\\

TabBERT introduces a hierarchical transformer that captures the hierarchy in sequence data, where each row is further divided into attributes. Twin Tower propose a supervised learning mechanism, employing an architecture with two parallel towers to capture attention across time and dimension, without learning intermediate embeddings. LUNA propose a framework that separately handles categorical and numerical values.  We provide an objective comparison of these three models in Table \ref{tab:compare}.

\begin{table*}[t]
\caption{Comparative Analysis of Transformer Architectures: Exploring Key Factors for Selecting the Optimal Architecture}
\label{tab:compare}
\scalebox{0.9}{
\begin{tabular}{|l|l|l|l|}
\hline
{\bf Model} & {\bf TabFormer (TabBERT)} & {\bf Gated Transformer (Twin Tower)} & {\bf LUNA} \\ \hline
Required Pre-training &  Yes& No& Yes\\ \hline
Pretraining method & Masked Language Modeling (MLM) & No & MLM (with regression loss) \\ \hline
Prediction level &Series (Multiple Rows) &Series (Multiple Rows) & Series (Multiple Rows)\\ \hline
Categorical values &As embeddings & As numerical values& As embeddings\\ \hline
Numerical values & As embeddings of categorical values &As numerical values &As numerical values \\ \hline
Architecture & Hierarchical Transformer&Twin tower & Hierarchical Transformer\\ \hline
Dataset Tested & 2 datasets,artificially generated&13 datasets & 1 public dataset\\ \hline
Prone to Overfitting & No& Yes& No\\ \hline
Training Time &High & Least& Medium\\ \hline
Data Pre-Processing &Data Quantization, Bulky Vocabulary &Easy Processing e.g. categorical encoding & Vocabulary creation for categorical values \\ \hline
Similar architectures & UniTTab, LUNA & CrossFormer, TARNet & Numerical reasoning based architecture\\ \hline

\end{tabular}}
\end{table*}

To understand the time complexity of the above models in terms of input size, we assume that each sequence consists of N rows:- \([r_1,r_2,....r_N]\) and each row consists of M attributes: \([k_1,k_2,...k_M]\).
For a transformer with a sequence of length N, due to multi-headed attention in encoder and decoder layers time complexity is \(\mathcal{O}(N^2)\). Since in hierarchical transformers such as TabBERT, for each row encoder attention is also computed at the attribute level the time complexity becomes:
\begin{equation}
    T(r_i,k_i)=\mathcal{O}(N^2*M^2)
\end{equation}
For Twin Tower, attention is parallel computed for different dimensions (across time and attributes). Then the outputs are combined using a gating channel \((f(O_1,O_2)=W_1*O_1+W_2*O_2)\). Hence the time complexity can be expressed as:
\begin{equation}
    T(r_i,k_i)=\mathcal{O}(N^2+M^2)
\end{equation}

Next, we discus various techniques for handling input data and training process.
\subsection{Data prepocessing}
Financial dataset comprises a combination of categorical and numerical (continuous) values. Various approaches can be employed to handle these values:

\begin{itemize}
\item \textbf{Converting numerical values to categorical values:} Pre-training transformers using a cross-entropy loss requires converting continuous or numerical values into categorical values. Techniques like binning/quantization or frequency encoding have been proposed for this conversion \cite{padhi2021tabular, luetto2023one}.
\item \textbf{Converting categorical values to numerical values:} For direct supervised training, categorical values can be treated as continuous values and directly passed to the transformer's embedding layer. Methods like binary encoding, one-hot encoding, or label encoding are used to convert categorical values into numerical values \cite{liu2021gated, chowdhury2022tarnet}.
\item \textbf{Treating numerical and categorical values separately:} A modified loss proposed by Han et al. \cite{han2022luna} treats numerical and categorical values separately during pre-training. This approach uses separate loss terms for regression (numerical) and cross-entropy (categorical) values, eliminating the need for value conversion and preventing information loss.
\end{itemize}

These different approaches provide flexibility in adapting the preprocessing step to the specific characteristics and requirements of the financial dataset.

\subsection{Training mechanisms}
Transformers in the language domain are typically trained through decoupled unsupervised pre-training and supervised fine-tuning due to the large vocabulary size. In the tabular domain, training can also be conducted using direct supervision or the decoupled approach similar to language models. Our exploration examines two training mechanisms:

\begin{itemize}
    \item \textbf{Decoupled pre-training and fine-tuning:} Separate pre-training and fine-tuning processes are used for tabular transformers \cite{huang2020tabtransformer,padhi2021tabular,han2022luna}. It is particularly useful when labeled data is limited or when the same corpus is utilized for multiple downstream tasks.
    
    \item \textbf{Direct supervised training:} For specific problems with ample labeled data, direct supervised training is employed. Transformer based approaches such as Liu et al. \cite{liu2021gated} and Zhang et al. \cite{zhang2023crossformer} use this approach.
\end{itemize}
TabBERT follows the approach of converting numerical values to categorical values and employs decoupled pre-training and fine-tuning. Twin Tower converts categorical values to numerical values and uses direct supervised training. LUNA modifies the loss function during pre-training to handle numerical values separately.

%% file: fraud.tex
\section{Synthetic data based study}
\label{sec:fraud}
This section describes our experimental evaluation on a public synthetic dataset.
\vspace{-0.3cm}
\subsection{Credit Fraud Prediction}
\begin{figure}[!ht]
    \centering
    \includegraphics[scale=0.25]{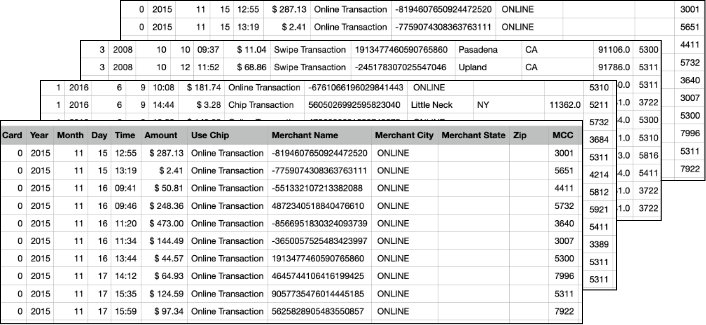}
    \caption{A snapshot of credit fraud prediction dataset}
    \label{fig:a}
\end{figure}

The fraud prediction dataset, \cite{padhi2021tabular}, comprises a diverse set of transactions generated artificially, involving various users. Each transaction is labeled as either fraudulent or non-fraudulent and includes multiple attributes, categorized as either categorical (e.g., 'use chip') or numerical (e.g., 'amount') (as illustrated in Figure \ref{fig:a}). The primary goal is to predict whether a user is engaged in fraudulent activity based on a sequence of their transactions. The adopted approach follows the original problem definition \cite{padhi2021tabular}, where, given a window of N consecutive transactions, if at least one is classified as fraudulent, the entire sequence is labeled as fraudulent.

It is crucial to note that the data distribution is highly imbalanced. Out of a total of 24 million transactions, only approximately 30,000 transactions belong to the fraudulent category. By applying the aforementioned labeling strategy with a window size of 10 (N=10), we obtain a mere 9,000 fraud sequences within a vast dataset of 2.4 million data points. We show the data distribution in Table \ref{tab:dataset}.

\begin{table}[!ht]
\centering
\caption{Data Distribution of the credit fraud problem}
\label{tab:dataset}
\begin{tabular}{|l|l|l|l|l|}
\hline
& \multicolumn{2}{c|}{\textbf{Transaction wise}} & \multicolumn{2}{c|}{\textbf{Sample wise}} \\
\hline
& \textbf{Train} & \textbf{Test} & \textbf{Train} & \textbf{Test} \\
\hline
Fraudulent & 23996 & 5761 & 7229 & 1797\\
Non-Fraudulent & 19485524 &  4871619 & 1943723 & 485940\\
\hline
\end{tabular}

\end{table}

\subsection{Performance comparison}
We focus on three models: TabBERT, Twin Tower, and LUNA. We also include the baseline scores of a standard vanilla architecture (the left tower in Twin Tower architecture discussed in the subsection \ref{subsec:Ablation}) 
trained via direct supervised training. We upsampled the data to equalize the frequency of fraudulent and non-fraudulent class using SMOTE \cite{rustogi2019swift} library. The evaluation metric is the binary F1 score on the fraudulent class. We do not use accuracy score as accuracy is not a suitable metric for evaluating tasks with imbalanced class ratios. We report precision, recall, and F1 score for these models in Table \ref{tab:F1}. 
\\
\begin{table}[!ht]
\centering
\caption{Performance comparison of different transformer architectures on credit fraud problem.}
\label{tab:F1}
\begin{tabular}{|l|l|l|l|}
\hline
{\bf Architecture}&{\bf Precision}&{\bf Recall}&{\bf $F_1$  score}\\ \hline
Vanilla&0.96&0.74&0.836\\
Twin Tower&0.95&0.76&0.844\\
TabBERT & 0.97 & 0.81&0.886 \\ 
LUNA &0.98&0.80&0.880 \\\hline
\end{tabular}
\end{table}
\vspace{-0.4cm}
\\

Furthermore, when considering the practical applicability of these models in an industry setting, factors such as training time and space requirements become important metrics for assessing their efficiency. To provide a more quantitative analysis of the training time for these models, we present the relevant information in Table \ref{tab:time}. We use Nvidia V100 16GB GPUs for our experiments. Refer Appendix \ref{appen:hpt-credit-fraud} for more details on the hyperparameters.
\begin{table}[!ht]
\centering
\caption{Training Time Comparison of different transformer architectures on credit fraud problem}
\label{tab:time}
\begin{tabular}{|l|l|c|}
\hline
{\bf Approach} & {\bf \# epochs} & {\bf Total time (Hrs)} \\
\hline
Vanilla & 20 & 3.6\\
\hline
Twin Tower & 50 & 27.5\\
\hline
TabBERT & 3 & 90 (pre-train) + 3 (fine-tuning) \\
\hline
LUNA & 3 & 62 (pre-train) + 3 (fine-tuning) \\
\hline
\end{tabular}
\end{table}

\vspace{-0.1cm}
\subsection{Ablation study - Time Dependence vs Dimension Dependence}
\label{subsec:Ablation}

During the experimentation with the Twin Tower architecture, a notable observation is made regarding the importance of attention weights across time steps and attribute dimensions. In the Twin Tower architecture, where the left tower focuses on interaction across time steps (referred to as the \textit{Vanilla} approach) and the right tower attends to interaction across features, we assess the significance of feature interactions in isolation. (Refer to \ref{appen:twin-tower} for more details on the architecture). An experiment is conducted by selectively masking one tower during training. The findings reveal that the isolated performance of the left tower closely resembles that of the complete Twin Tower model. However, the isolated performance of the right tower is poor, suggesting that, in the dataset used, temporal information holds more importance compared to interactions across features. To summarize the outcomes of this study, we present the results in Table \ref{tab:twin tower}. 

\vspace{-0.1cm}
\begin{table}[!ht]
\centering
\caption{Performance scores with attention captured along different dimensions}
\vspace{-0.2cm}
\label{tab:twin tower}
\scalebox{0.9}{
\begin{tabular}{|l|l|l|l|}
\hline
{\bf Method}&{\bf Precision}&{\bf Recall}&{\bf $F_1$ score}\\ \hline
Time interaction tower (Vanilla) & 0.96 & 0.74 & 0.84\\
Feature interaction tower & 0.38 & 0.19 & 0.26\\
\hline
\end{tabular}
}
\vspace{-0.2cm}
\end{table}

%% file: cdss.tex
\section{Industry data based study}
\label{sec:cdss}
We next report the experimental investigation on industry data.
Credit default prediction is crucial for robust risk modeling, enabling the identification of customers who might fail to repay debts, thereby preventing significant financial losses. The task involves predicting a binary variable based on a customer's performance within an 18-month window after their latest credit card statement. A default event is recorded if the customer fails to pay the due amount within 120 days after the statement date.\footnote{\url{https://www.kaggle.com/competitions/amex-default-prediction/}} 

Table \ref{tab:datadist} provides details on the data distribution for the train, val \& test set respectively. The dataset comprises approximately 190 features for each record, extracted from various thematic categories such as Delinquency, Spend, Payment, Balance, and Risk-based variables. This extensive customer information over a temporal nature poses a complex challenge due to the sheer volume of data. 

Dealing with missing values in industry datasets is a notable challenge, about 122 features have missing values. During data analysis we observed that there are two main broader categories of missing values: i) features missing few values for a single month across 13 months, ii) for about 15 features more than 50\% of the total records were missing. In literature there has been multiple methods recommended for handling imputation for missing values, using mean, mode, median, max, min etc., \cite{donders2006gentle} or dropping the feature all together \cite{graham2003methods} depending on the nature of variables. As such transformers have shown to handle missing values and noise, in prior research \cite{fang2020time}. Hence, in our explorations we went with replacing missing values with zero (0), for the model to learn signals to capture the representation of missing values and noise in data better using attention mechanism. However, this is an interesting area of exploration for future extension to identify trade-offs with different imputations while working with large feature sets and that too for industry datasets on the scale of 5-10M records.

\vspace{-0.1cm}
\begin{table}[!ht]
\centering
\vspace{-0.2cm}
\caption{Data distribution}
\vspace{-0.2cm}
\label{tab:datadist}
\scalebox{0.9}{
\begin{tabular}{|l|l|l|}
\hline
{\bf Data Type}&{\bf Records}&{\bf Customers}\\ \hline
Train Data & 5.8M & 460K\\
Val Data & 5.7M& 470k \\
Test Data & 5.6M& 460k\\
\hline
\end{tabular}
}
\vspace{-0.1cm}
\end{table}

Given that tree-based models are commonly employed for modeling financial datasets, the baseline model for this study utilizes a LightGBM-based approach. The results of various transformer models and the baseline scores are presented in Table \ref{tab:cdss-results}. The evaluation metric $M$ for this problem is the mean of two rank-ordering measures: the Normalized Gini Coefficient ($G$) and the default rate captured at 4\% ($D$). The default rate captured at 4\% represents the percentage of positive labels (defaults) captured within the highest-ranked 4\% of predictions and serves as a Sensitivity/Recall statistic.
\vspace{-0.2cm}
\begin{equation}
        M = 0.5*(G+D)
\end{equation}

The metric $M$ has a maximum value of 1.0. Both sub-metrics $G$ and $D$ assign a weight of 20 to negative labels to account for downsampling. Additionally, F1 score evaluation is provided for better comparison with synthetic data studies. Given the similarities in training mechanisms and performance between LUNA and TabBERT observed in the credit fraud problem, the analysis focuses exclusively on TabBERT for the credit default problem. TabBERT was pre-trained on AWS p3.8xlarge, equipped with 4 V100 GPUs, maximum memory of 244GB. The entire process of pre-training and fine-tuning TabBERT took approximately 2.5 days, while Twin Tower and LightGBM took 2.5 hrs and 4 hrs, respectively. More details on hyperparameters can be referred in section \ref{appen:hpt-credit-default}.

\begin{table}[!ht]
\caption{Transformer results on industry based dataset}
\vspace{-0.2cm}
\label{tab:cdss-results}
\begin{tabular}{|l|l|l|l|l|}
\hline
{\bf Architecture}&{\bf Metric $M$}&{\bf Gini}&{\bf Capture Rate}&{\bf $F_1$  Score}\\ \hline
LightGBM&79.29&91.87&66.72 & 0.783\\
Vanilla&79.43&91.95&66.91& 0.792\\
Twin Tower&\textbf{79.86}&\textbf{92.17}&\textbf{67.56}& \textbf{0.795}\\ 
TabBERT & 71.70& 88.56& 54.89& 0.708\\\hline
\end{tabular}
\vspace{-0.3cm}
\end{table}

{\bf Comparative Analysis:} The TabBERT model, which performed well on synthetic datasets, demonstrates significant under performance on an industrial dataset. The model struggles to effectively learn embeddings for the extensive vocabulary present in financial datasets. Converting numerical features into categorical ones results in information loss, diminishing discriminative power, and relying on arbitrary cutoff points due to data quantization mechanism. This approach overlooks the similarity between closely related numbers assigned to different categories, which is crucial for problems sensitive to small input variations \cite{gholami2021survey}. In contrast, the Twin-Tower approach, which employs a direct supervised training approach, proves to be effective and efficient for this use case, closely followed by the Vanilla approach.

\section{Experiments with Regression tasks}
\label{subsect:generalization}
The findings and lessons explored in this work for the different transformer modeling approaches, are not limited to these two classification datasets but is also applicable to other classification datasets on structured data as well as for different regression tasks. We share some findings and analysis on the explorations conducted on regression problem for synthetic and industry scale datasets.

\subsection{Synthetic Dataset}
For comparing performance of these different transformer approached on a regression task, we experimented with Pollution prediction dataset \cite{misc_beijing_multi-site_air-quality_data_501}. Task is to predict PM2.5 and PM10 air concentration for 12 monitoring sites, each containing around 35k entries (rows). This is a commonly used public dataset for regression prediction on a multi-variate time series based data and have been used as a benchmark for evaluating transformer models for regression tasks \cite{padhi2021tabular,luetto2023one}. Dataset has about 400k data points, every row has 11 fields with both numerical and categorical values. 
For a detailed description of the data, please refer \cite{liang2015assessing}. Data has missing values, and we replace missing data with zero (0), in our experiments. Table \ref{tab:regression_public} present the results of different transformer models.

\begin{table}[!ht]
\centering
\caption{Performance comparison of different transformer architectures on public dataset for regression problem. $^{+}$ as per the results reported in \cite{padhi2021tabular}}
\label{tab:regression_public}
\begin{tabular}{|l|l|l|l|}
\hline
{\bf Architecture}&{\bf RMSE}\\ \hline
Vanilla& 53.6\\
Twin Tower& 54.2\\
TabBERT$^{+}$ & {\bf 32.8}\\ \hline
\end{tabular}
\end{table}
\vspace{-0.3cm}
\subsection{Industry dataset}

We experiment with a very common problem in finance industry know as spend prediction \cite{jin2018prediction}. The problem is quite known but have not been explored in-depth due to lack of proper industry scale real datasets. Given consumer data records comprising of different features such as their credit card spend behaviors for last 12 months or so, credit bureau scores, etc., task is to predict their future spend. This model forms a foundational model for marketing and other incentives offered to a consumer. This data consisted of 200k customer data with 1 record per month from 13 months data for each customer so total data points amounting to 2.6M records. Each data point has 148 feature attributes \footnote{Due to privacy reasons, this dataset cannot be released.}. We used 70\% of the data for training and 15\% for validation and 15\% for testing. Table \ref{tab:regressions_industry} presents the results of different transformers for spend prediction.
 
\begin{table}[!ht]
\caption{Different transformer results on industry based dataset for regression}
\vspace{-0.4cm}
\label{tab:regressions_industry}
\begin{tabular}{|l|l|l|l||l|}
\hline
{\bf Architecture}&{\bf RMSE}\\ \hline
LightGBM&25618\\
Vanilla&25370\\
Twin Tower& {\bf 24471}\\ 
TabBERT & 66126\\\hline
\end{tabular}
\vspace{-0.3cm}
\end{table}

In this work, we've noticed a similar trend in how regression models perform compared to the classification models we discussed earlier in this paper. Direct supervised techniques like Twin Tower work work well with large and complex datasets while decoupled pre-training and fine-tuning techniques like TabBERT have some limitations dealing with them. We find that these techniques can be used for different types of tasks involving tabular understanding. Importantly, our findings can be applied to a wide range of table-related tasks, not limited to specific tasks we tested.

%% file: lessons-learnt.tex
\section{Lessons Learned}
\label{sec:lessons}
We present the key insights gained from the experimentation discussed in Sections \ref{sec:methodology}, \ref{sec:fraud}, and \ref{sec:cdss}.

\begin{itemize} 
    \vspace{0.1cm}
    \item {\bf Upsampling effect: }
    Fraud prediction and finance-related datasets often show class imbalance but at the same time use unweighted evaluation metrics. For fraud prediction using a synthetic dataset, we employed the SMOTE library, which generates artificial data points using a nearest neighbor algorithm. Notably, we observe significant performance improvements with SMOTE-based upsampling with the techniques such as Twin Tower as shown in Table \ref{tab:twin tower upsample}. Without upsampling F1 score is 0.608 and with upsampling it increased to 0.844.  
    \begin{table}[!ht]
    \centering
    \caption{Twin Tower results for credit fraud prediction with and without upsampling}
    \label{tab:twin tower upsample}
    \scalebox{0.9}{
    \begin{tabular}{|l|l|l|l|}
    \hline
    {\bf Method}&{\bf Precision}&{\bf Recall}&{\bf $F_1$ score}\\ \hline
    Without Upsampling& 0.74 & 0.51 & 0.608\\
    With Upsampling & 0.95 & 0.76 & 0.844\\
    \hline
    \end{tabular}
    }
    \end{table}
    \item {\bf Architectures for tabular series: } 
    The study focuses on architectures for modeling sequential tabular data, emphasizing attention across both attribute dimensions and time steps. The ablation study in Section \ref{sec:fraud} demonstrates the crucial importance of attention across time steps for the model's performance. The findings suggest that simplifying the architectures to prioritize attention across time steps provides a practical solution for working with large datasets under resource and space constraints.
    \vspace{0.1cm}
    \item {\bf Hyper Parameter Optimization: } 
    Selecting the right parameters is crucial for achieving optimal model performance. Parameters such as the number of attention heads, hidden layer dimensions, learning rate, optimizer, etc., can be effectively chosen using libraries such as RayTune \cite{liaw2018tune}. The study showcases performance enhancements in credit default prediction achieved by fine-tuning hyperparameters, as detailed in Table \ref{tab:HPT}.
    
    \begin{table}[!ht]
    \centering
    \vspace{-0.1cm}
    \caption{Performance scores with and without hyperparameter optimization on credit default prediction problem}
    \vspace{-0.3cm}
    \label{tab:HPT}
    \scalebox{0.85}{
    \begin{tabular}{|l|l|l|l|l|}
    \hline
    {\bf Architecture}&{\bf Metric $M$}&{\bf Gini}&{\bf Capture Rate}&{\bf $F_1$ Score}\\ \hline
    Twin Tower&79.53&92.07&66.99&0.792\\
    Twin Tower (with HPT)&\textbf{79.86}&\textbf{92.17}&\textbf{67.56}& 0.795\\ 
    \hline
    \end{tabular}
    }
    \end{table}
    \vspace{0.2cm}
    
    \item {\bf Infrastructure and compute resources :}
    As discussed in Section \ref{sec:cdss}, training models such as TabBERT for credit default prediction poses challenges due to memory constraints. Industry datasets often present large scale and high dimensionality. Therefore, simpler architectures for models are preferred, as they can offer comparable performance to pre-training-based models while consuming fewer resources.
    \end{itemize}
 

%% file: conclusion.tex
\section{Conclusion}
\label{sec:conclusion}
This study provides a thorough analysis of transformer techniques for financial tabular series, covering input processing, architecture, training mechanisms, and upsampling. Evaluation on both artificial and real-world industry datasets highlights the Twin Tower model as a suitable choice for industry-scale datasets, offering reduced space requirements and competitive F1 scores. The findings emphasize the importance of considering essential factors in model selection. The study aims to advance transformer applications in financial analysis, offering guidance for researchers and enhancing practical usability. Future work aims to improve interpretability and address challenges with missing and noisy data.

%% file: appendix.tex
\appendix
\renewcommand{\thesection}{A}
\section{Appendix}

\subsection{Hyperparameters for Credit Fraud Prediction}
\label{appen:hpt-credit-fraud}
For the credit fraud problem, we share the list of hyperparameters (refer Table \ref{tab:parameters}) for our experiments described in section \ref{sec:fraud}.
\begin{table}[!ht]
\centering
\caption{Hyperparameter values for Credit Fraud Problem}
\label{tab:parameters}
\begin{tabular}{|l|l|l|l|}
\hline
{\bf Architecture} & {\bf TabBERT} & {\bf Twin Tower} & {\bf LUNA} \\
\hline
Learning Rate & 5e-5 & 4.35e-5 & 5e-5 \\
\hline
Optimiser & Adam & Adam & Adam\\
\hline
Dropout & 0.1 & 0.134 & 0.1\\
\hline
Attention heads & 12 & 8 & 12\\
\hline
Hidden units & 768 & 256 & 768 \\
\hline
Window size & 10 & 10 & 10 \\
\hline
Stride & 5 & 1 & 10\\
\hline
Batch size & 8 & 256 & 8\\
\hline
MLM Probability & 0.15 & -- & 0.15\\
\hline
\end{tabular}
\vspace{-0.3cm}
\end{table}

\subsection{Hyperparameters for Industry Based Study - Credit Default Prediction}
\label{appen:hpt-credit-default}
For the credit default problem, we share the list of hyperparameters (refer Table  \ref{tab:paramters-gbm} and \ref{tab:parameters-cdss} ) for our experiments described in section \ref{sec:cdss}.

\begin{table}[!ht]
    \centering
    \caption{LightGBM Hyperparameter values for Credit Default Prediction}
    \label{tab:paramters-gbm}
    \begin{tabular}{|c|c|} \hline
         {\bf LightGBM model} & {\bf Values} \\ \hline
         No. of leaves & 100\\ \hline
         Min data in leaf & 2\\ \hline
         No. of boosting rounds & 2000 \\ \hline
         Early stopping rounds & 50 \\ \hline
         Learning Rate & 0.01 \\ \hline
         Seed & 42 \\ \hline
         Max depth &  default (-1) \\ \hline
    \end{tabular}
\end{table}

\begin{table}[!ht]
\centering
\caption{Hyperparameter values for Credit Default Prediction}
\label{tab:parameters-cdss}
\begin{tabular}{|l|l|l|}
\hline
{\bf Architecture} & {\bf TabBERT} & {\bf Twin Tower}  \\
\hline
Learning Rate & 0.01 & 1e-4 \\
\hline
Optimiser & Adam & Adam \\
\hline
Dropout & 0.1 & 0.1 \\
\hline
Attention heads & 12 & 12\\
\hline
Seed & 9 & 42 \\
\hline
Hidden units & 768 & 512 \\
\hline
Window size & 12 & 12 \\
\hline
Batch size & 16 & 512\\
\hline
MLM Probability & 0.15 & NA \\
\hline
\end{tabular}
\vspace{-0.3cm}
\end{table}
Additionally, we present a loss convergence plot during the training of TabBERT for the credit default problem. As mentioned in section \ref{sec:cdss}, our analysis reveals that TabBERT's performance is sub-optimal compared to direct supervised training methods such as Twin Tower when applied to real industry-scale datasets.
\begin{figure} [!ht]
    \centering
    \includegraphics[scale=0.3]{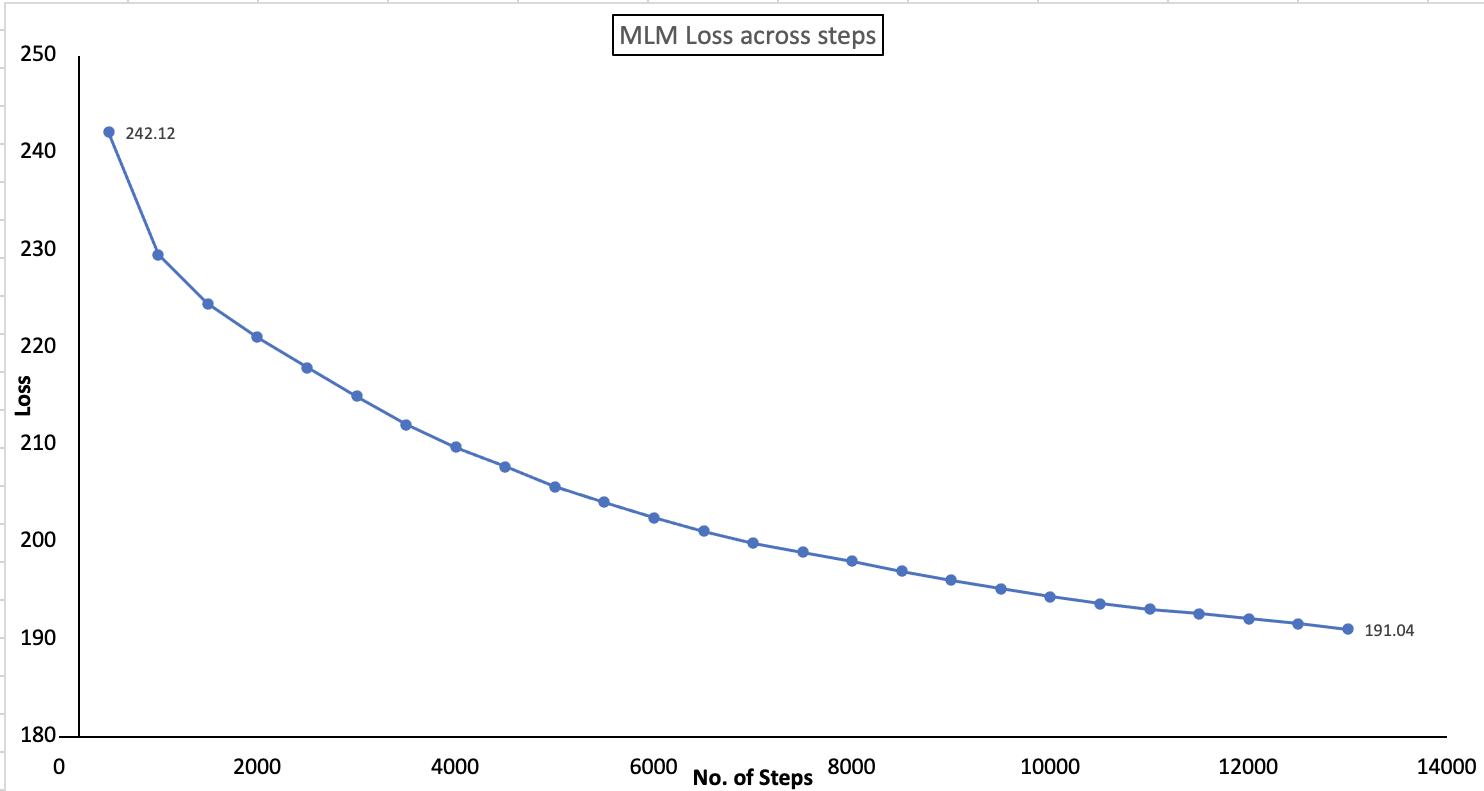}
    \caption{MLM loss convergence for TabBERT pre-training}
    \label{fig:mlmloss}
   \vspace{-0.3cm}
\end{figure}

\subsection{Twin Tower architecture of the Gated Transformer}
\label{appen:twin-tower}
As discussed in section \ref{sec:methodology}, the Gated Transformer by \cite{liu2021gated} consists of two parallel transformer blocks. The left block captures attention across different time steps whereas the right transformer captures attention across attribute dimension. The output from both the blocks are combined using a gated channel. We present a study to independently assess the contribution of each tower in combined performance of the model via ablation study presented in subsection \ref{subsec:Ablation}.
\begin{figure}[!ht]
    \centering
    \includegraphics[scale=0.7]{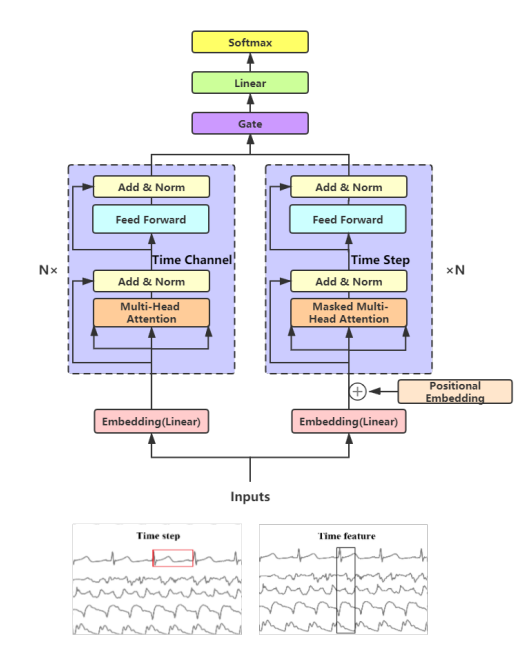}
    \caption{Image describing the architecture of Gated Transformer (referred from \citet{liu2021gated}). The left tower captures attention across time step while right tower captures attention across attribute dimension.}
    \label{fig:Gated}
\end{figure}

%% file: main.bbl

\begin{thebibliography}{28}


\ifx \showCODEN    \undefined \def \showCODEN     #1{\unskip}     \fi
\ifx \showDOI      \undefined \def \showDOI       #1{#1}\fi
\ifx \showISBNx    \undefined \def \showISBNx     #1{\unskip}     \fi
\ifx \showISBNxiii \undefined \def \showISBNxiii  #1{\unskip}     \fi
\ifx \showISSN     \undefined \def \showISSN      #1{\unskip}     \fi
\ifx \showLCCN     \undefined \def \showLCCN      #1{\unskip}     \fi
\ifx \shownote     \undefined \def \shownote      #1{#1}          \fi
\ifx \showarticletitle \undefined \def \showarticletitle #1{#1}   \fi
\ifx \showURL      \undefined \def \showURL       {\relax}        \fi
\providecommand\bibfield[2]{#2}
\providecommand\bibinfo[2]{#2}
\providecommand\natexlab[1]{#1}
\providecommand\showeprint[2][]{arXiv:#2}

\bibitem[Alvi et~al\mbox{.}(2021)]%
        {alvi2021deep}
\bibfield{author}{\bibinfo{person}{Redwan~Hasif Alvi}, \bibinfo{person}{Md~Habibur Rahman}, \bibinfo{person}{Adib Al~Shaeed Khan}, {and} \bibinfo{person}{Rashedur~M Rahman}.} \bibinfo{year}{2021}\natexlab{}.
\newblock \showarticletitle{Deep learning approach on tabular data to predict early-onset neonatal sepsis}.
\newblock \bibinfo{journal}{\emph{Journal of Information and Telecommunication}} \bibinfo{volume}{5}, \bibinfo{number}{2} (\bibinfo{year}{2021}), \bibinfo{pages}{226--246}.
\newblock


\bibitem[Chen(2019)]%
        {misc_beijing_multi-site_air-quality_data_501}
\bibfield{author}{\bibinfo{person}{Song Chen}.} \bibinfo{year}{2019}\natexlab{}.
\newblock \bibinfo{title}{{Beijing Multi-Site Air-Quality Data}}.
\newblock \bibinfo{howpublished}{UCI Machine Learning Repository}.
\newblock
\newblock
\shownote{{DOI}: https://doi.org/10.24432/C5RK5G}.


\bibitem[Chen and Guestrin(2016)]%
        {chen2016xgboost}
\bibfield{author}{\bibinfo{person}{Tianqi Chen} {and} \bibinfo{person}{Carlos Guestrin}.} \bibinfo{year}{2016}\natexlab{}.
\newblock \bibinfo{booktitle}{\emph{XGBoost: {A} Scalable Tree Boosting System}}.
\newblock \bibinfo{publisher}{{ACM}}. 785--794 pages.
\newblock
\urldef\tempurl%
\url{https://doi.org/10.1145/2939672.2939785}
\showDOI{\tempurl}


\bibitem[Chowdhury et~al\mbox{.}(2022)]%
        {chowdhury2022tarnet}
\bibfield{author}{\bibinfo{person}{Ranak~Roy Chowdhury}, \bibinfo{person}{Xiyuan Zhang}, \bibinfo{person}{Jingbo Shang}, \bibinfo{person}{Rajesh~K Gupta}, {and} \bibinfo{person}{Dezhi Hong}.} \bibinfo{year}{2022}\natexlab{}.
\newblock \showarticletitle{TARNet: Task-Aware Reconstruction for Time-Series Transformer}.
\newblock  (\bibinfo{year}{2022}), \bibinfo{pages}{14--18}.
\newblock


\bibitem[Donders et~al\mbox{.}(2006)]%
        {donders2006gentle}
\bibfield{author}{\bibinfo{person}{A~Rogier~T Donders}, \bibinfo{person}{Geert~JMG Van Der~Heijden}, \bibinfo{person}{Theo Stijnen}, {and} \bibinfo{person}{Karel~GM Moons}.} \bibinfo{year}{2006}\natexlab{}.
\newblock \showarticletitle{A gentle introduction to imputation of missing values}.
\newblock \bibinfo{journal}{\emph{Journal of clinical epidemiology}} \bibinfo{volume}{59}, \bibinfo{number}{10} (\bibinfo{year}{2006}), \bibinfo{pages}{1087--1091}.
\newblock


\bibitem[Dorogush et~al\mbox{.}(2018)]%
        {dorogush2018catboost}
\bibfield{author}{\bibinfo{person}{Anna~Veronika Dorogush}, \bibinfo{person}{Vasily Ershov}, {and} \bibinfo{person}{Andrey Gulin}.} \bibinfo{year}{2018}\natexlab{}.
\newblock \showarticletitle{CatBoost: gradient boosting with categorical features support}.
\newblock \bibinfo{journal}{\emph{ArXiv preprint}}  \bibinfo{volume}{abs/1810.11363}.
\newblock
\urldef\tempurl%
\url{https://arxiv.org/abs/1810.11363}
\showURL{%
\tempurl}


\bibitem[Du et~al\mbox{.}(2021)]%
        {du2021tabularnet}
\bibfield{author}{\bibinfo{person}{Lun Du}, \bibinfo{person}{Fei Gao}, \bibinfo{person}{Xu Chen}, \bibinfo{person}{Ran Jia}, \bibinfo{person}{Junshan Wang}, \bibinfo{person}{Jiang Zhang}, \bibinfo{person}{Shi Han}, {and} \bibinfo{person}{Dongmei Zhang}.} \bibinfo{year}{2021}\natexlab{}.
\newblock \bibinfo{booktitle}{\emph{TabularNet: A neural network architecture for understanding semantic structures of tabular data}}.
\newblock 322--331 pages.
\newblock


\bibitem[Fang and Wang(2020)]%
        {fang2020time}
\bibfield{author}{\bibinfo{person}{Chenguang Fang} {and} \bibinfo{person}{Chen Wang}.} \bibinfo{year}{2020}\natexlab{}.
\newblock \showarticletitle{Time series data imputation: A survey on deep learning approaches}.
\newblock \bibinfo{journal}{\emph{arXiv preprint arXiv:2011.11347}} (\bibinfo{year}{2020}).
\newblock


\bibitem[Gholami et~al\mbox{.}(2021)]%
        {gholami2021survey}
\bibfield{author}{\bibinfo{person}{Amir Gholami}, \bibinfo{person}{Sehoon Kim}, \bibinfo{person}{Zhen Dong}, \bibinfo{person}{Zhewei Yao}, \bibinfo{person}{Michael~W Mahoney}, {and} \bibinfo{person}{Kurt Keutzer}.} \bibinfo{year}{2021}\natexlab{}.
\newblock \showarticletitle{A survey of quantization methods for efficient neural network inference}.
\newblock \bibinfo{journal}{\emph{ArXiv preprint}}  \bibinfo{volume}{abs/2103.13630} (\bibinfo{year}{2021}).
\newblock
\urldef\tempurl%
\url{https://arxiv.org/abs/2103.13630}
\showURL{%
\tempurl}


\bibitem[Graham et~al\mbox{.}(2003)]%
        {graham2003methods}
\bibfield{author}{\bibinfo{person}{John~W Graham}, \bibinfo{person}{Patricio~E Cumsille}, {and} \bibinfo{person}{Elvira Elek-Fisk}.} \bibinfo{year}{2003}\natexlab{}.
\newblock \showarticletitle{Methods for handling missing data}.
\newblock \bibinfo{journal}{\emph{Handbook of psychology}} (\bibinfo{year}{2003}), \bibinfo{pages}{87--114}.
\newblock


\bibitem[Guo et~al\mbox{.}(2017)]%
        {guo2017deepfm}
\bibfield{author}{\bibinfo{person}{Huifeng Guo}, \bibinfo{person}{Ruiming Tang}, \bibinfo{person}{Yunming Ye}, \bibinfo{person}{Zhenguo Li}, {and} \bibinfo{person}{Xiuqiang He}.} \bibinfo{year}{2017}\natexlab{}.
\newblock \showarticletitle{DeepFM: {A} Factorization-Machine based Neural Network for {CTR} Prediction}. In \bibinfo{booktitle}{\emph{Proceedings of the Twenty-Sixth International Joint Conference on Artificial Intelligence, {IJCAI} 2017, Melbourne, Australia, August 19-25, 2017}}, \bibfield{editor}{\bibinfo{person}{Carles Sierra}} (Ed.). \bibinfo{publisher}{ijcai.org}, \bibinfo{pages}{1725--1731}.
\newblock
\urldef\tempurl%
\url{https://doi.org/10.24963/ijcai.2017/239}
\showDOI{\tempurl}


\bibitem[Han et~al\mbox{.}(2022)]%
        {han2022luna}
\bibfield{author}{\bibinfo{person}{Hongwei Han}, \bibinfo{person}{Jialiang Xu}, \bibinfo{person}{Mengyu Zhou}, \bibinfo{person}{Yijia Shao}, \bibinfo{person}{Shi Han}, {and} \bibinfo{person}{Dongmei Zhang}.} \bibinfo{year}{2022}\natexlab{}.
\newblock \showarticletitle{LUNA: Language Understanding with Number Augmentations on Transformers via Number Plugins and Pre-training}.
\newblock \bibinfo{journal}{\emph{ArXiv preprint}}  \bibinfo{volume}{abs/2212.02691} (\bibinfo{year}{2022}).
\newblock
\urldef\tempurl%
\url{https://arxiv.org/abs/2212.02691}
\showURL{%
\tempurl}


\bibitem[Hazimeh et~al\mbox{.}(2020)]%
        {hazimeh2020tree}
\bibfield{author}{\bibinfo{person}{Hussein Hazimeh}, \bibinfo{person}{Natalia Ponomareva}, \bibinfo{person}{Petros Mol}, \bibinfo{person}{Zhenyu Tan}, {and} \bibinfo{person}{Rahul Mazumder}.} \bibinfo{year}{2020}\natexlab{}.
\newblock \showarticletitle{The Tree Ensemble Layer: Differentiability meets Conditional Computation}. In \bibinfo{booktitle}{\emph{Proceedings of the 37th International Conference on Machine Learning, {ICML} 2020, 13-18 July 2020, Virtual Event}} \emph{(\bibinfo{series}{Proceedings of Machine Learning Research}, Vol.~\bibinfo{volume}{119})}. \bibinfo{publisher}{{PMLR}}, \bibinfo{pages}{4138--4148}.
\newblock
\urldef\tempurl%
\url{http://proceedings.mlr.press/v119/hazimeh20a.html}
\showURL{%
\tempurl}


\bibitem[Huang et~al\mbox{.}(2019)]%
        {huang2019dsanet}
\bibfield{author}{\bibinfo{person}{Siteng Huang}, \bibinfo{person}{Donglin Wang}, \bibinfo{person}{Xuehan Wu}, {and} \bibinfo{person}{Ao Tang}.} \bibinfo{year}{2019}\natexlab{}.
\newblock \showarticletitle{DSANet: Dual Self-Attention Network for Multivariate Time Series Forecasting}. In \bibinfo{booktitle}{\emph{Proceedings of the 28th {ACM} International Conference on Information and Knowledge Management, {CIKM} 2019, Beijing, China, November 3-7, 2019}}, \bibfield{editor}{\bibinfo{person}{Wenwu Zhu}, \bibinfo{person}{Dacheng Tao}, \bibinfo{person}{Xueqi Cheng}, \bibinfo{person}{Peng Cui}, \bibinfo{person}{Elke~A. Rundensteiner}, \bibinfo{person}{David Carmel}, \bibinfo{person}{Qi~He}, {and} \bibinfo{person}{Jeffrey~Xu Yu}} (Eds.). \bibinfo{publisher}{{ACM}}, \bibinfo{pages}{2129--2132}.
\newblock
\urldef\tempurl%
\url{https://doi.org/10.1145/3357384.3358132}
\showDOI{\tempurl}


\bibitem[Huang et~al\mbox{.}(2020)]%
        {huang2020tabtransformer}
\bibfield{author}{\bibinfo{person}{Xin Huang}, \bibinfo{person}{Ashish Khetan}, \bibinfo{person}{Milan Cvitkovic}, {and} \bibinfo{person}{Zohar Karnin}.} \bibinfo{year}{2020}\natexlab{}.
\newblock \showarticletitle{Tabtransformer: Tabular data modeling using contextual embeddings}.
\newblock \bibinfo{journal}{\emph{ArXiv preprint}}  \bibinfo{volume}{abs/2012.06678} (\bibinfo{year}{2020}).
\newblock
\urldef\tempurl%
\url{https://arxiv.org/abs/2012.06678}
\showURL{%
\tempurl}


\bibitem[Jin et~al\mbox{.}(2018)]%
        {jin2018prediction}
\bibfield{author}{\bibinfo{person}{Liyin Jin}, \bibinfo{person}{Jaimie~W Lien}, {and} \bibinfo{person}{Junji Xiao}.} \bibinfo{year}{2018}\natexlab{}.
\newblock \showarticletitle{Prediction and Learning About Credit Card Spending}.
\newblock \bibinfo{journal}{\emph{Available at SSRN 3172869}} (\bibinfo{year}{2018}).
\newblock


\bibitem[Kadra et~al\mbox{.}(2021)]%
        {kadra2021well}
\bibfield{author}{\bibinfo{person}{Arlind Kadra}, \bibinfo{person}{Marius Lindauer}, \bibinfo{person}{Frank Hutter}, {and} \bibinfo{person}{Josif Grabocka}.} \bibinfo{year}{2021}\natexlab{}.
\newblock \showarticletitle{Well-tuned Simple Nets Excel on Tabular Datasets}. In \bibinfo{booktitle}{\emph{Advances in Neural Information Processing Systems 34: Annual Conference on Neural Information Processing Systems 2021, NeurIPS 2021, December 6-14, 2021, virtual}}, \bibfield{editor}{\bibinfo{person}{Marc'Aurelio Ranzato}, \bibinfo{person}{Alina Beygelzimer}, \bibinfo{person}{Yann~N. Dauphin}, \bibinfo{person}{Percy Liang}, {and} \bibinfo{person}{Jennifer~Wortman Vaughan}} (Eds.). \bibinfo{pages}{23928--23941}.
\newblock
\urldef\tempurl%
\url{https://proceedings.neurips.cc/paper/2021/hash/c902b497eb972281fb5b4e206db38ee6-Abstract.html}
\showURL{%
\tempurl}


\bibitem[Ke et~al\mbox{.}(2017)]%
        {ke2017lightgbm}
\bibfield{author}{\bibinfo{person}{Guolin Ke}, \bibinfo{person}{Qi Meng}, \bibinfo{person}{Thomas Finley}, \bibinfo{person}{Taifeng Wang}, \bibinfo{person}{Wei Chen}, \bibinfo{person}{Weidong Ma}, \bibinfo{person}{Qiwei Ye}, {and} \bibinfo{person}{Tie{-}Yan Liu}.} \bibinfo{year}{2017}\natexlab{}.
\newblock \showarticletitle{LightGBM: {A} Highly Efficient Gradient Boosting Decision Tree}. In \bibinfo{booktitle}{\emph{Advances in Neural Information Processing Systems 30: Annual Conference on Neural Information Processing Systems 2017, December 4-9, 2017, Long Beach, CA, {USA}}}, \bibfield{editor}{\bibinfo{person}{Isabelle Guyon}, \bibinfo{person}{Ulrike von Luxburg}, \bibinfo{person}{Samy Bengio}, \bibinfo{person}{Hanna~M. Wallach}, \bibinfo{person}{Rob Fergus}, \bibinfo{person}{S.~V.~N. Vishwanathan}, {and} \bibinfo{person}{Roman Garnett}} (Eds.). \bibinfo{pages}{3146--3154}.
\newblock
\urldef\tempurl%
\url{https://proceedings.neurips.cc/paper/2017/hash/6449f44a102fde848669bdd9eb6b76fa-Abstract.html}
\showURL{%
\tempurl}


\bibitem[Ke et~al\mbox{.}(2018)]%
        {ke2018tabnn}
\bibfield{author}{\bibinfo{person}{Guolin Ke}, \bibinfo{person}{Jia Zhang}, \bibinfo{person}{Zhenhui Xu}, \bibinfo{person}{Jiang Bian}, {and} \bibinfo{person}{Tie-Yan Liu}.} \bibinfo{year}{2018}\natexlab{}.
\newblock \showarticletitle{TabNN: A universal neural network solution for tabular data}.
\newblock  (\bibinfo{year}{2018}).
\newblock


\bibitem[Liang et~al\mbox{.}(2015)]%
        {liang2015assessing}
\bibfield{author}{\bibinfo{person}{Xuan Liang}, \bibinfo{person}{Tao Zou}, \bibinfo{person}{Bin Guo}, \bibinfo{person}{Shuo Li}, \bibinfo{person}{Haozhe Zhang}, \bibinfo{person}{Shuyi Zhang}, \bibinfo{person}{Hui Huang}, {and} \bibinfo{person}{Song~Xi Chen}.} \bibinfo{year}{2015}\natexlab{}.
\newblock \showarticletitle{Assessing Beijing's PM2. 5 pollution: severity, weather impact, APEC and winter heating}.
\newblock \bibinfo{journal}{\emph{Proceedings of the Royal Society A: Mathematical, Physical and Engineering Sciences}} \bibinfo{volume}{471}, \bibinfo{number}{2182} (\bibinfo{year}{2015}), \bibinfo{pages}{20150257}.
\newblock


\bibitem[Liaw et~al\mbox{.}(2018)]%
        {liaw2018tune}
\bibfield{author}{\bibinfo{person}{Richard Liaw}, \bibinfo{person}{Eric Liang}, \bibinfo{person}{Robert Nishihara}, \bibinfo{person}{Philipp Moritz}, \bibinfo{person}{Joseph~E Gonzalez}, {and} \bibinfo{person}{Ion Stoica}.} \bibinfo{year}{2018}\natexlab{}.
\newblock \showarticletitle{Tune: A research platform for distributed model selection and training}.
\newblock \bibinfo{journal}{\emph{ArXiv preprint}}  \bibinfo{volume}{abs/1807.05118} (\bibinfo{year}{2018}).
\newblock
\urldef\tempurl%
\url{https://arxiv.org/abs/1807.05118}
\showURL{%
\tempurl}


\bibitem[Liu et~al\mbox{.}(2021)]%
        {liu2021gated}
\bibfield{author}{\bibinfo{person}{Minghao Liu}, \bibinfo{person}{Shengqi Ren}, \bibinfo{person}{Siyuan Ma}, \bibinfo{person}{Jiahui Jiao}, \bibinfo{person}{Yizhou Chen}, \bibinfo{person}{Zhiguang Wang}, {and} \bibinfo{person}{Wei Song}.} \bibinfo{year}{2021}\natexlab{}.
\newblock \showarticletitle{Gated transformer networks for multivariate time series classification}.
\newblock \bibinfo{journal}{\emph{ArXiv preprint}}  \bibinfo{volume}{abs/2103.14438} (\bibinfo{year}{2021}).
\newblock
\urldef\tempurl%
\url{https://arxiv.org/abs/2103.14438}
\showURL{%
\tempurl}


\bibitem[Luetto et~al\mbox{.}(2023)]%
        {luetto2023one}
\bibfield{author}{\bibinfo{person}{Simone Luetto}, \bibinfo{person}{Fabrizio Garuti}, \bibinfo{person}{Enver Sangineto}, \bibinfo{person}{Lorenzo Forni}, {and} \bibinfo{person}{Rita Cucchiara}.} \bibinfo{year}{2023}\natexlab{}.
\newblock \showarticletitle{One Transformer for All Time Series: Representing and Training with Time-Dependent Heterogeneous Tabular Data}.
\newblock \bibinfo{journal}{\emph{ArXiv preprint}}  \bibinfo{volume}{abs/2302.06375} (\bibinfo{year}{2023}).
\newblock
\urldef\tempurl%
\url{https://arxiv.org/abs/2302.06375}
\showURL{%
\tempurl}


\bibitem[Padhi et~al\mbox{.}(2021)]%
        {padhi2021tabular}
\bibfield{author}{\bibinfo{person}{Inkit Padhi}, \bibinfo{person}{Yair Schiff}, \bibinfo{person}{Igor Melnyk}, \bibinfo{person}{Mattia Rigotti}, \bibinfo{person}{Youssef Mroueh}, \bibinfo{person}{Pierre Dognin}, \bibinfo{person}{Jerret Ross}, \bibinfo{person}{Ravi Nair}, {and} \bibinfo{person}{Erik Altman}.} \bibinfo{year}{2021}\natexlab{}.
\newblock \showarticletitle{Tabular transformers for modeling multivariate time series}. In \bibinfo{booktitle}{\emph{ICASSP 2021-2021 IEEE International Conference on Acoustics, Speech and Signal Processing (ICASSP)}}. IEEE, \bibinfo{pages}{3565--3569}.
\newblock


\bibitem[Popov et~al\mbox{.}(2020)]%
        {popov2019neural}
\bibfield{author}{\bibinfo{person}{Sergei Popov}, \bibinfo{person}{Stanislav Morozov}, {and} \bibinfo{person}{Artem Babenko}.} \bibinfo{year}{2020}\natexlab{}.
\newblock \showarticletitle{Neural Oblivious Decision Ensembles for Deep Learning on Tabular Data}. In \bibinfo{booktitle}{\emph{8th International Conference on Learning Representations, {ICLR} 2020, Addis Ababa, Ethiopia, April 26-30, 2020}}. \bibinfo{publisher}{OpenReview.net}.
\newblock
\urldef\tempurl%
\url{https://openreview.net/forum?id=r1eiu2VtwH}
\showURL{%
\tempurl}


\bibitem[Rustogi and Prasad(2019)]%
        {rustogi2019swift}
\bibfield{author}{\bibinfo{person}{Rishabh Rustogi} {and} \bibinfo{person}{Ayush Prasad}.} \bibinfo{year}{2019}\natexlab{}.
\newblock \showarticletitle{Swift imbalance data classification using SMOTE and extreme learning machine}. In \bibinfo{booktitle}{\emph{2019 International Conference on Computational Intelligence in Data Science (ICCIDS)}}. IEEE, \bibinfo{pages}{1--6}.
\newblock


\bibitem[Somepalli et~al\mbox{.}(2021)]%
        {somepalli2021saint}
\bibfield{author}{\bibinfo{person}{Gowthami Somepalli}, \bibinfo{person}{Micah Goldblum}, \bibinfo{person}{Avi Schwarzschild}, \bibinfo{person}{C~Bayan Bruss}, {and} \bibinfo{person}{Tom Goldstein}.} \bibinfo{year}{2021}\natexlab{}.
\newblock \showarticletitle{Saint: Improved neural networks for tabular data via row attention and contrastive pre-training}.
\newblock \bibinfo{journal}{\emph{ArXiv preprint}}  \bibinfo{volume}{abs/2106.01342} (\bibinfo{year}{2021}).
\newblock
\urldef\tempurl%
\url{https://arxiv.org/abs/2106.01342}
\showURL{%
\tempurl}


\bibitem[Zhang and Yan(2023)]%
        {zhang2023crossformer}
\bibfield{author}{\bibinfo{person}{Yunhao Zhang} {and} \bibinfo{person}{Junchi Yan}.} \bibinfo{year}{2023}\natexlab{}.
\newblock \showarticletitle{Crossformer: Transformer utilizing cross-dimension dependency for multivariate time series forecasting}. In \bibinfo{booktitle}{\emph{The Eleventh International Conference on Learning Representations}}.
\newblock


\end{thebibliography}
